\documentclass[11pt,a4paper]{article}
\usepackage[hyperref]{conll-2019}
\usepackage{times}
\usepackage{latexsym}

\usepackage{url}

\usepackage[english]{babel}
\usepackage[utf8]{inputenc}
\usepackage{amsmath}
\usepackage{amsthm} % for theoremstyle
\usepackage{amsfonts}
\usepackage{graphicx}
\usepackage{booktabs}
\usepackage{hyperref}
\usepackage{natbib}
\usepackage{pgfplots}
\usepackage{balance}
\pgfplotsset{compat=1.7}
\usepackage{subfig}
\usepackage{caption}
\usepackage{enumitem}

\aclfinalcopy % Uncomment this line for the final submission
\def\aclpaperid{290} %  Enter the acl Paper ID here

\theoremstyle{definition}
\newtheorem{definition}{Notation}[section]

\newcommand\mainword{target word}

\DeclareMathOperator*{\argmin}{arg\,min}

\newcommand\norm[1]{\left\lVert#1\right\rVert}

\title{Generating Timelines by Modeling Semantic Change}

\author{Guy D. Rosin\textsuperscript{1}, Kira Radinsky\textsuperscript{1,2} \\
         \textsuperscript{1}Technion -- Israel Institute of Technology, Haifa, Israel \\
         \textsuperscript{2}eBay Research, Israel \\
         {\tt \{guyrosin,kirar\}@cs.technion.ac.il}
}

\date{}

\begin{document}

\maketitle

\begin{abstract}
Though languages can evolve slowly, they can also react strongly to dramatic world events. By studying the connection between words and events, it is possible to identify which events change our vocabulary and in what way. In this work, we tackle the task of creating timelines---records of historical `turning points', represented by either words or events, to understand the dynamics of a \mainword{}.
Our approach identifies these points by leveraging both static and time-varying word embeddings to measure the influence of words and events. In addition to quantifying changes, we show how our technique can help isolate semantic changes. Our qualitative and quantitative evaluations show that we are able to capture this semantic change and event influence.

%The field of language dynamics has been studied in recent years, but still little is known about the dynamics of lexical evolution.
%Language evolution is contingent upon historical factors, and specifically upon historical events.
%By studying relations between words and events over time, we can learn how events affect language.
%In this paper, we introduce the task of historical etymology generation, i.e., identifying historical turning points in the lifetime of words, and explaining them using events.
%We use both static and temporal word embeddings to measure changes in words and learn the influences of events on particular words.
%Specifically, major societal transformations, as well as catastrophic events such as wars lead to an increased change in words semantics. 
%In addition, we introduce the task of identifying whether an event affected a particular word's semantics.
%Our qualitative and quantitative tests show that our methods succeed in capturing semantic change, as well as events influence on words.
\end{abstract}

\maketitle

\section{Introduction}
\label{sec:introduction}

Languages respond to world events in many ways. New words, phrases, and named entities are created, new senses may develop, and valences may change. Various approaches support the study of historical linguistics (e.g., comparative linguistics, etymology, etc.). In this work, we focus on a specific process for tracking the progression of meaning over time in sense, semantics, and in relation to other words and concepts. By leveraging changing relationships in temporal corpora, we demonstrate a way of `embedding' words and world events. Observing changes in this embedding allows us to construct timelines that support the study of evolving languages.

The timeline of scientific and technical discoveries, for example, can drive the emergence of new word senses as these discoveries are `named'. Take the word ``cell'' which evolved from its $12^{th}$ century meaning (a small room or chamber) to a new sense in the $17^{th}$ century (a basic unit of an organism) to the $19^{th}$ century meaning (an electric battery) and most recently to a shorthand for a mobile phone\footnote{\url{https://www.etymonline.com/word/cell}}. Critically, the \textit{dominant senses} of a word vary over time as some meanings become less commonly used while others gain in popularity. This dynamic need not be driven only by the addition of certain senses. The prevalence of a hyponym, for example, may also drive a change in the ranking of senses. The word `disaster' may call to mind very different things depending on the latest \textit{type} of disaster. Thus, the word may evoke `nuclear disaster' in a reader in 2011 (e.g., driven by the \textit{Fukushima} incident). However, in 2012 the `storm' sense may be more salient (e.g., driven by \textit{Superstorm Sandy}). 

Evolution of senses is but one way a language can evolve. Broader semantic changes can also occur. For example, the valence of the word may move or even flip (e.g., \textit{terrific} or \textit{bully}). Of particular interest to us are those changes that are more immediate and precipitated by key world events. For example, a war may lead certain terms to take on a negative connotation as a country or people become the `enemy'. Large collections of text from a given period can capture all of these language changes as reflected by evolving context. By mining this text, our goal is to support the study of evolving languages.

Etymological studies allow us to understand the origin of words and changes in meaning~\cite{alinei1995thirty}. This work produces not only an accounting of change but also an explanation of the social, scientific, or other world events that drive language shifts. 
Conventional production of etymological analysis often requires a detailed and laborious manual close-reading of historical texts \cite{geeraerts1997diachronic}. 
%The goal of our work is to support the construction of these \etymology{}s automatically. 
By applying computational methods, our focus is on detecting semantic changes of words and events and producing possible explanations from real-world drivers.

\begin{figure}
\centering
\includegraphics[width=0.47\textwidth]{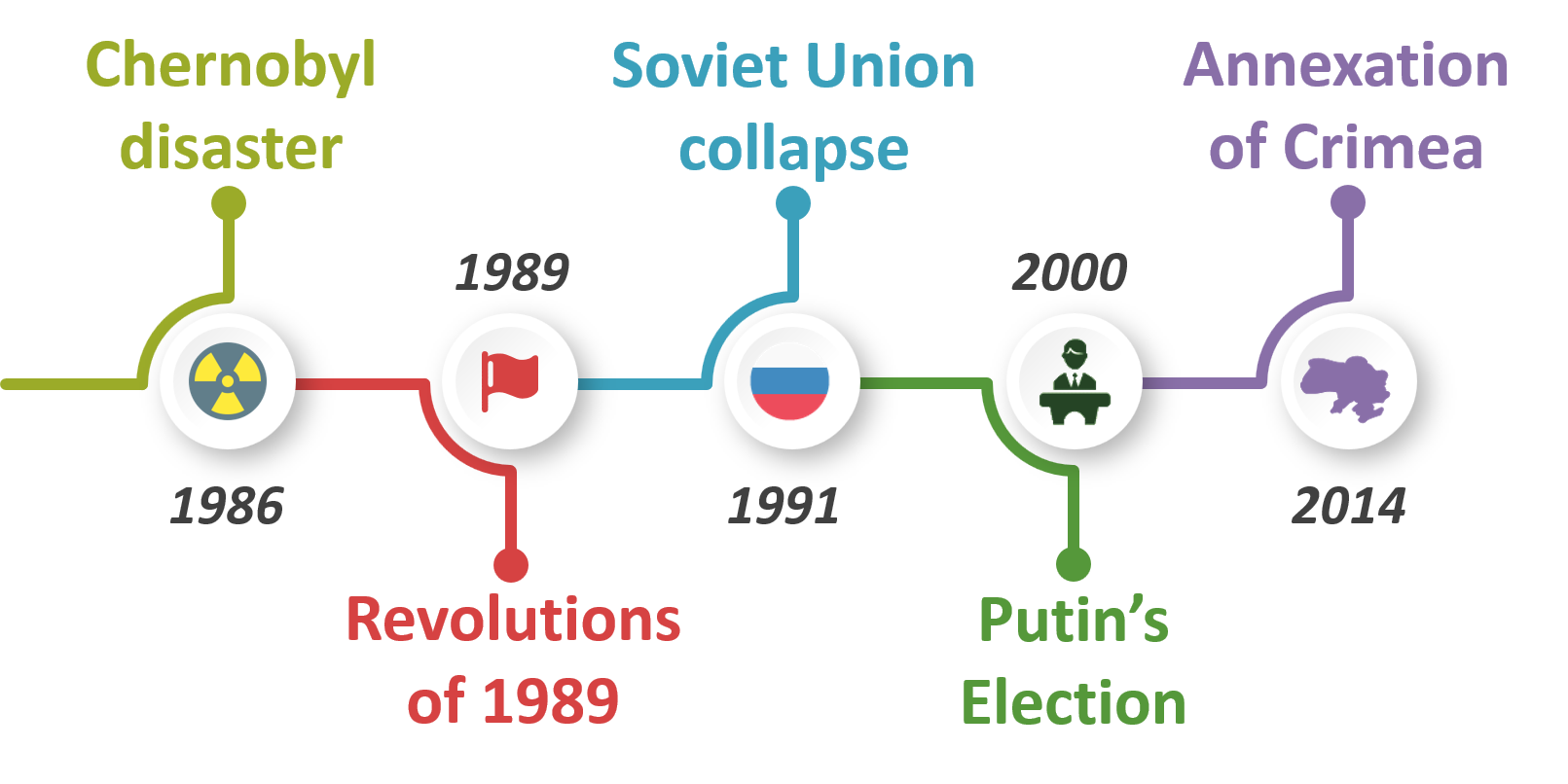}
\caption{\label{fig:timeline_example}A timeline generated by our framework for \textit{Russia}.}
%\vspace{-3mm}
\end{figure}

\textbf{Problem Definition}: In this work, we study the problem of timeline generation for a word or phrase. We use the term `word' throughout the paper for convenience.
Given a timeline of a word, a researcher should be able to understand the word's dynamics, i.e., the changes the word underwent over time. A timeline is defined as a sequence of time points and their descriptors (i.e., explanations of the changes the word underwent at that time point). A good timeline is such that enables the researcher to gain a better understanding of the word and its history \citep{althoff2015timemachine}. It contains time points of significant changes, with relevant explanations of the changes, and with a minimal number of missing or redundant information.

% building blocks of a timeline
We define several building blocks for constructing timelines. The first is the identification of time points during which the \mainword{} underwent significant semantic change (we refer to those as \textit{Turning Points}, see Section~\ref{sec:year_detection}). 
Second, we consider identifying associated descriptors of those changes.
These descriptors are associated with a word's change at a particular time and can serve to explain its dynamics.

We experiment with two types of descriptors. The first involves \textit{words} associated with the \mainword{} or affected by it (Section~\ref{sec:word_descriptors}). The above `cell' and `disaster' examples can serve as examples of timelines with word descriptors. The second type is \textit{events} (Section~\ref{sec:event_descriptors}). One can explain changes the \mainword{} underwent based on significant world events. As an example, consider the timeline generated by our framework for the word ``Russia'' (Figure~\ref{fig:timeline_example}). 

% how to identify the descriptors
To identify events that are strongly associated with the change, we utilize time-varying language embeddings on both static snapshots (e.g., Wikipedia) and historical texts (35 years of the \textit{New York Times}), allowing us to capture both syntactic and semantic variation of words (Section~\ref{sec:embeddings}).
We develop a mechanism for simultaneously embedding words and events in the same space (Section~\ref{sec:projected_embeddings}).
We present several methods to leverage those embeddings for key historical events detection by evaluating the distance between words and events (Sections \ref{sec:similarity_events}, \ref{sec:classifier}).
% how to represent events and words

\begin{figure}
\centering
\includegraphics[width=0.48\textwidth]{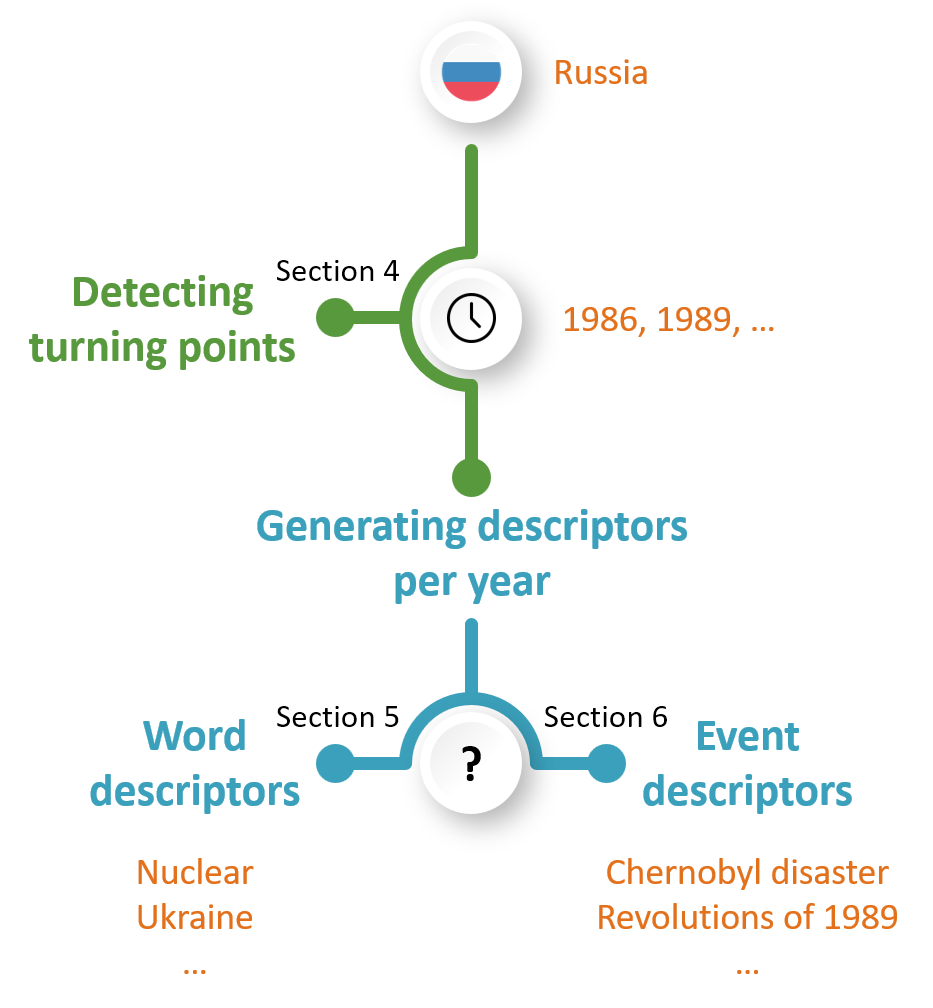}
\caption{\label{fig:timeline_flow}Flow diagram of timeline generation. The two basic building blocks are detecting turning points and generating descriptors, where the descriptors can be either words or events.}
\end{figure}

Figure~\ref{fig:timeline_flow} presents the flow of the paper through an example. Consider the word `Russia'. First, we identify its turning points (Section~\ref{sec:year_detection}), and then generate descriptors -- either words (Section~\ref{sec:word_descriptors}) or events (Section~\ref{sec:event_descriptors}) -- to construct its timeline.
% See the flow diagram in Figure~\ref{fig:timeline_flow} for a summary of the process of timeline generation.
We contribute several algorithms for identifying significant events leveraging various types of embeddings, including a supervised learning approach.\footnote{Code and data available at \url{https://github.com/guyrosin/generating_timelines}}

\section{Related Work}
\label{sec:related_work}

\textbf{Semantic Change}:
Most work on language evolution has focused on identifying semantic drifts and word meaning changes (see~\citet{DiachronicSurvey2018} for a recent survey). Various approaches have pursued the task of detecting changes in word meaning~\cite{Sagi:2009:SDA,Mitra:2014,Wijaya:2011:USC,Mihalcea:2012,Popescu:2013,Jatowt:2014:FAS,KenterAndRijke:2015:CIKM,hamilton2016diachronic,azarbonyad2017words}. 
Specific approaches include: dynamic embedding models using a probabilistic Bayesian version of Word2Vec \cite{bamler2017dynamic}, pointwise mutual information (PMI) \cite{yao2018dynamic}, and exponential family embeddings \cite{rudolph2018dynamic}. Relatedness over time between words has also been studied. \citet{Radinsky:2011} showed that words that co-occur in history have a stronger relation, \citet{rosin2017learning} introduced the supervised task of temporal semantic relatedness, and \citet{orlikowski2018learning} studied diachronic analogies.
In our work, we focus on the world events behind the semantic changes---and isolate those events that co-occur with significant language change.

\textbf{Change Detection for Semantic Shift Analysis}: Detecting major changes involves detecting continuous peaks in time series. \citet{kulkarni2015statistically} and \citet{basile2016diachronic} offered a mean shift model, and \citet{rosin2017learning} used a threshold-based method for this task. We utilize the latter approach as it is simpler and more computationally effective.

% \textbf{Etymology:}
% Etymology is most commonly defined as the study of the origin of words and the way in which their meanings changed throughout history. Many other definitions exist \cite{alinei1995thirty}. %, e.g., ``Whether etymology studies the origin of words is not sure; what it does study is the history of words, and in particular their phonetic and semantic changes.''
% In this work, we focus on the historical changes of words and study \textit{\etymology{}s} -- a record of historical turning points in the lifetime of words. 
% %Specifically, we are interested in explaining changes in word semantics.
% Work in the related field of historical lexicology was done by \citet{geeraerts1997diachronic}, where he demonstrated the value of the prototype model of meaning. In our work, we attempt to perform those tasks in an automated manner.

\textbf{Timeline Generation}:
Past work focused on generating timelines by leveraging information retrieval methods. Examples include the use of Facebook data~\citep{graus2013yourhistory}, Twitter \citep{li2014timeline}, and Wikipedia~\citep{tuan2011cate} to generate context-aware timelines by ranking related entities by co-occurrence with the main timeline entity.
\citet{althoff2015timemachine} created timelines by mining a knowledge base, based on submodular optimization and web-co-occurrence statistics. \citet{shahaf2010connecting} generated a chain of events connecting two news articles.
Our work differs from the prior work in several ways. First, we consider the semantic changes a word undergoes and detect the events that influenced them. We study several word embeddings to measure change and relatedness between words and events over time and construct a timeline.

\section{Event and Temporal Word Embedding}
\label{sec:embeddings}
We consider several methods to represent events and 
words. These are used to identify semantic changes and generate timeline descriptors. To capture both words and events in the same space we consider \textit{global embeddings}, which are created upon the English Wikipedia (see Section~\ref{sec:implementation_details}).

We also utilize the work of \citet{rosin2017learning} for \textit{temporal word embeddings}. Timeline construction requires modeling changes in words and events over time. When looking at a specific word, we wish to focus on its relevant meaning at a particular time.
Thus, the \textit{temporal word embeddings} are created using data from a large temporal corpus. Specifically, we leverage the New York Times (NYT) archive. The embeddings are generated for every time period (i.e., year) and enable us to investigate how words meanings and relatedness between words change over time (see Section~\ref{sec:implementation_details}).

Finally, in order to compare vectors of the same word in different, independently-trained, vector space models, we align every pair of models using Orthogonal Procrustes \citep{hamilton2016diachronic}.

We use the following notations throughout the paper:
\begin{definition}
$v_w$ is the vector representation of a word $w$.
\end{definition}
\begin{definition}
$v^t_w$ is the vector representation of a word $w$ during time $t$.
\end{definition}
\begin{definition}
${NN}_k(w)$ is the set of k-nearest neighbors (kNN) of a word $w$.
\end{definition}
\begin{definition}
${NN}_k^t(w)$ is the set of k-nearest neighbors (kNN) of a word $w$ during time $t$.
\end{definition}
\begin{definition}
$cos$ is cosine similarity, which we use as a similarity function between embeddings.
\end{definition}

\section{Timeline Turning Points}
\label{sec:year_detection}
A timeline is composed of time points that identify the changes a word underwent. We refer to those as \emph{Turning Points}, and experiment with several methods to identify them. Formally, let $w$ be a \mainword{}, and $t$ be a time point. Each method approximates the probability $d_t(w)$ of $t$ to be a turning point of $w$.
The turning points are then selected by performing peak detection~\cite{rosin2017learning} on the series of $d_t(w)$ for every $t$.
We experiment with two methodologies for turning point detection, leveraging the embeddings we introduce in Sections~\ref{sec:embeddings}, \ref{sec:projected_embeddings}:

\textbf{(1) Neighborhood}: Changes in the neighborhood of a word over time can be used to capture semantic changes of the word.
This method measures the difference between the similar words sets of $w$ between two consecutive years. Formally:
$
d_t(w) = 1 - \frac{\big| {NN}_k^t(w) \cap {NN}_k^{t-1}(w) \big|} {k}
$,
where ${NN}_k^t(w)$ is the set of k-nearest neighbors (kNN) of $w$ during time $t$.

\textbf{(2) EmbeddingSimilarity}: Employing word embeddings, we can also look at the change in the embedding vectors of $w$:
$
d_t(w) = 1 - {cos}\big(v^t_w, v^{t-1}_w\big)
$

\section{Word Descriptors}
\label{sec:word_descriptors}
The second building block of a timeline is \textit{explaining} what triggered the semantic changes. We refer to such explanations as \textit{descriptors}.
One can explain semantic changes with words---significant associated words or terms that correlate to the \mainword{}'s meaning drift. Letting $w$ be the \mainword{}, and $t$ be a time point, we are looking for words that became closer to $w$ in vector space during $t$. Specifically, we look at the nearest neighbors of $w$ at times $t$ and $t-1$, and denote the set of descriptors by $D_t$:
$
D_t(w) = {NN}_k^t(w) \setminus {NN}_k^{t-1}(w)
$.
%where ${NN}_k^t(w)$ is the set of k-nearest neighbors (kNN) of the word $w$ during time $t$.
For example, using this method for `Russia' results in \textit{Soviet Union} and \textit{Soviet} for 1989, when the Soviet Union was dissolving, and \textit{Ukraine}, \textit{Kyrgyzstan}, and \textit{Latvia} for 1990, when these countries attempted to gain independence from the Union.
%Table~\ref{tab:etymology_entities_comparison} contains an \etymology{} for Russia that was generated using this method, where the etymological descriptors are words. All the words that appear in the timeline were indeed significant for Russia: in 1989 the Soviet Union was dissolving; in 1990 and 1991 the four countries -- Ukraine, Kyrgyzstan, Latvia, and Romania -- attempted to gain independence from the Soviet Union. Such descriptors often provide a clue about the nature of the change, but may not directly connect to known events.

% \begin{table}
%   \tiny
%   \centering
%   \begin{tabular}{lll}
%     \toprule
%     Year & Words & Events \\
%     \midrule
%     1989 & Soviet Union, Soviet & Revolutions of 1989 \\
%     1990 & Ukraine, Kyrgyzstan, Latvia & UNESCO Designation of Moscow Kremlin \\
%     1991 & Romania & Dissolution of the Soviet Union, START I \\
%     \bottomrule
%   \end{tabular}
%   \caption{Example \etymology{}s for ``Russia'', with descriptors represented by words (in the center) and by events (in the right column).}
%   \label{tab:etymology_entities_comparison}
% \end{table}

\section{Event Descriptors}
\label{sec:event_descriptors}
As an alternative for word descriptors, we consider event descriptors.
To generate these, given a \mainword{} $w$, our task is to identify its change in time $t$ by a set of significant events $E$ that likely affected $w$. 
In this section, we first describe the embeddings we use (Section~\ref{sec:projected_embeddings}) and then present several methods for detecting significant events (Sections~\ref{sec:similarity_events}, \ref{sec:classifier}).

\subsection{Projected Embedding for Events}
\label{sec:projected_embeddings}
Temporal word embeddings (Section~\ref{sec:embeddings}) are created for every time period. They enable us to investigate how word meaning and relatedness between words change over time. However, as it may take some time until an event's name is determined and referred to in newspapers, the paper's text may not have meaningful embeddings for those events. For example, the name ``World War I'' was used only after WWII started. As a result, we are not able to compare events and words. %in a specific time.

To address this problem, we leverage the global embeddings (Section~\ref{sec:embeddings}). Since Wikipedia articles typically contain balanced descriptions of events, they can be a proper basis for event embeddings. Recall that the way these embeddings are created enables creating a common latent space for \textit{both} words and concepts (and specifically, events), allowing us to compare both at the same time.
Our solution involves projecting the global model on each temporal one. This way, we create a joint vector space for words and events, which represents a specific time period. We refer to these embeddings as ``\textit{Projected Embeddings}''.

We assume that most words’ meanings do not change over time and learn a transformation of one embedding space onto another, minimizing the distance between pairs of points.
%For each year, we take the global Wikipedia model and project it on the specific year's space from the NYT, using the method described below.
Let us define $W_{wiki} \in \mathbb{R}^{|V_{wiki}| \times d_{wiki}}$ as the matrix of embeddings learned from Wikipedia, where $d_{wiki}$ is the embedding size and $|V_{wiki}|$ is the vocabulary size of Wikipedia. Similarly, $W_{nyt}^{(t)} \in \mathbb{R}^{|V_{nyt}^{(t)}| \times d_{nyt}^{(t)}}$ is the matrix of embeddings learned from the NYT at time $t$, where $d_{nyt}^{(t)}$ is the embedding size and $|V_{nyt}^{(t)}|$ is the vocabulary size of the NYT at time $t$. We seek a matrix $W^{(t)} \in \mathbb{R}^{|V_{wiki}| \times d_{nyt}^{(t)}}$ that will contain the transformation of $W_{wiki}$ to $W_{nyt}^{(t)}$ for time $t$.
By making an additional simplifying assumption that the vector spaces are equivalent under a linear transformation, we are able to find $W^{(t)}$ by optimizing the following linear regression model:
\begin{equation}
\small
\widehat{T}^{(t)} = \argmin_{T} \sum_{w_i \in V_{wiki} \cap V_{nyt}^{(t)}} \norm{W_{wiki}(w_i)T - W_{nyt}^{(t)}(w_i)}_2^2
\end{equation}
where $\widehat{T}^{(t)} \in \mathbb{R}^{d_{wiki} \times d_{nyt}^{(t)}}$.
We then obtain the projected matrix $W^{(t)} = W_{wiki} \widehat{T}^{(t)}$.
Similar methods were used in the field of temporal semantics to align embeddings of different time periods to a unified coordinate system \citep{szymanski2017temporal,kulkarni2015statistically}.

\subsection{Similarity-Based Event Detection}
\label{sec:similarity_events}
We hypothesize that the events closest in vector space to a word should be the most significant to its timeline. We experiment with two score functions that are based on semantic similarity. The descriptors are chosen as the top-scoring events.

\textbf{ByWord}:
Given a \mainword{} $w$, we look for its closest events in a specific time.
We define a score function of an event $e$: $score(e) = cos(v_w, v_e)$
where $cos$ is cosine similarity, and $v_w$ and $v_e$ are the respective embeddings of $w$ and $e$.

\textbf{ByKNN}: 
We wish to extend the ``impact circle'' of the event, as events sometimes do not affect a word directly but through other words.
We look for events that are closest not only to the \mainword{} but also to its neighbors.
We calculate the following score for every possible event $e$:

\vspace{-1mm}
\begin{small}
$$score(e) = avg(\{ cos(v_n, v_e) : n  \in \{w\} \cup {NN}_k(w) \})$$
\end{small}
\vspace{-5mm}

\subsection{Supervised Event Detection}
\label{sec:classifier}
The previous methods discuss only the semantic similarity of words and events, where we are actually interested in the probability of events to affect words. There are often multiple possible events that can act as explanations. 
Consider the following example. In 2010, several events related to Russia happened: New START (nuclear arms reduction treaty between the USA and Russia) was signed, there was a Winter Olympics, and the ROKS Cheonan (a South Korean warship) sunk. All relate to Russia, but only one appears to indicate a meaningful change to it---the New START event. Identifying the right events is a highly challenging task, as most usually all the candidate events are impactful events, related to the \mainword{}, and have an impact on various words due to their significance.
As another example, Nelson Mandela's death is semantically the closest event to the word `Clinton' during 2013, based on our projected embeddings. Though it is surely a powerful event and one that is related to many world leaders, it is hard to describe it as a turning point for either Bill or Hillary Clinton.
Therefore, we attempt to \textit{learn} which are the most relevant events for a given word.
We present a classifier that receives an event and a word, and outputs the probability this event affected that word.
This classifier functions as a predictor of the probability of an event $e$ to cause a semantic change of a word $w$. %It can be used as a final re-ranking step of each of the methods introduced before (Sections \ref{sec:method_events_by_word} and \ref{sec:method_events_by_knn}).

\subsubsection*{Training Data}
\label{sec:classifier_dataset}
To create training data for the classifier, we can use any embedding model that embeds both events and words, namely the global or the projected embeddings (Section~\ref{sec:embeddings} and~\ref{sec:projected_embeddings}). Given an event, we find terms affected by it and terms that are not.

\textbf{Affected Terms}: We limit the set of possible affected terms by an event and consider semantically similar terms to the event (we consider cosine similarity $>$ 0.3). A term is considered to be affected by an event if the term's meaning changed during the time of the event, and did not change in the year before the event.
%Note that we can’t limit the change after the event, because it might affect language even many years after it actually happened.

\textbf{Unaffected Terms}: Given an event, we sample terms from its semantically similar terms that were changed in the years \textit{after} the event, and were not changed during the year it happened. This way, we try to capture terms that are related to the event but were changed due to other reasons.

\subsubsection*{Machine Learning Approach}
\label{sec:classifier_features}
We consider several supervised machine learning approaches, experimenting with random forest, SVM, neural networks, etc. We also devise several features leveraged by our classifiers:

%\vspace{-1mm}
\begin{itemize}[itemsep=-.8ex]
\item $v_e$ and $v_w$, i.e., the embeddings of the event $e$ and the word $w$. Any embedding model (Sections~\ref{sec:embeddings}, \ref{sec:projected_embeddings}) can be used.
\item The semantic similarity between $v_e$ and $v_w$.%: $similarity(w, e)$., measured using one of the embeddings we described in Section~\ref{sec:representing_entities}.
\item Categories of the event $e$, taken from DBpedia and represented using bag-of-words. Each event is associated with one or more categories in DBpedia (e.g., social event, sports, military conflict). The bag-of-words vector comprises the top 150 categories. 
\item Features that indicate the event's popularity: number of internal and external links in $e$'s Wikipedia page, pageviews count of $e$'s Wikipedia page\footnote{We used Wikimedia Foundation's API to get pageviews of a single month (specifically October 2017).}. Intuitively, a popular event might be impactful. 
\end{itemize}

\section{Experimental Setup}
\label{sec:experimental_setup}
We briefly describe our dataset and embeddings before focusing on the evaluation.

\subsection{Implementation Details}
\label{sec:implementation_details}
\textbf{Embeddings}: The global embeddings were created based on the Wikipedia dump of May 2016, using Word2Vec's skip-gram with negative sampling, with a window size of 10. 
Following \citet{sherkat2017vector}, we perform a pre-processing step necessary for the embedding process to capture Wikipedia concepts and not just words: each inner link in the text (i.e., a link to a Wikipedia article) is replaced by an ID, so that every link to the same page is replaced by this page's ID.
After filtering low-frequency words and concepts, we find 3.2M unique embeddings, of which 1.7M are concepts.

For constructing temporal embeddings, we used the NYT archive\footnote{\url{http://spiderbites.nytimes.com/}}, with articles from 1981 to 2016 (9GB of text in total). For each year of content, we created embeddings using Word2Vec’s skip-gram with negative sampling, with a window size of 5 and a dimensionality of 140, using the Gensim library \citep{gensim}. We filtered out words with less than 50 occurrences.% during that year. 
%The one-year interval was chosen empirically, as we need it to be large enough to have enough data, but not too large, otherwise we would not be able to isolate the effects of a single event.
%We also tried using embeddings for every 5-years period, but that brought inaccurate performance.

\textbf{Events Data}: We used DBpedia and Wikipedia as sources for events.
First, we mined entities from DBpedia whose type is `event' (i.e., \texttt{yago/Event100029378}, \texttt{Ontology/Event}), and that have an associated Wikipedia page and associated year of occurrence. Second, we mined 50K events from Wikipedia's monthly events pages\footnote{For example, \url{https://en.wikipedia.org/wiki/Category:February_1992_events}} and retained the 30K events corresponding to our focus time period of 1981 to 2016. In this work, we focus on large, significant events, since these have the most potential to affect language \citep{chieu2004query}. Thus, we retained events with over 6000 monthly views (in October 2017) and over 15 external references. Our final dataset contains 1233 events, most related to armed conflicts, politics, disasters, and sports.

\textbf{Classifier Dataset}: To construct a ``ground truth'' dataset for our classifier, we find pairs of events (from the events dataset) and their affected and unaffected terms, as described in Section~\ref{sec:classifier_dataset}. For each event, we use either the global or projected embeddings to find 20 affected terms (there may be less, depending on the sensitivity of the change detection algorithm) and 20 unaffected terms. The dataset contains 21K pairs in total.

\subsection{Experimental Methodology}
\label{sec:manual_timeline_evaluation}
We perform both qualitative and quantitative evaluations. First, we conduct user studies to capture the utility of our algorithms as they might be used in practice (e.g., in search engine results).
%In addition, we evaluate the performance of the events classifier and the contribution of the projection method.
Twenty evaluators participated using real events data (Section~\ref{sec:implementation_details}).
We select a set of \mainword{}s based on two criteria: popularity---so that most evaluators would know them and preferably parts of their history; and the number of significant related events---so that meaningful timelines would be produced. In practice, we selected 30 \mainword{}s based on popularity (as expressed in the number of page views of the corresponding Wikipedia page).
For a given \mainword{}, evaluators were presented with several timelines created by our algorithms and baseline methods (see below). 
While we evaluated both word and event descriptors for timelines, our evaluation is focused on events, as they proved to be more meaningful and interesting to the evaluators (see Section~\ref{sec:timeline_eval}).
Each timeline was accompanied by detailed descriptions and references. Our evaluators were asked to indicate whether an event was correct (`true') in its placement in the timeline and whether it was likely to have an impact on the \mainword{}. See Appendix A for a screenshot of the questionnaire used in the evaluation. Overall timeline quality was evaluated as described below.

\subsubsection*{Evaluation Metrics}
\label{sec:evaluation_metrics}
Each timeline is evaluated using the following metrics (timelines with word descriptors are evaluated similarly---replacing `event' with `word'):

\textbf{Accuracy}: Fraction of events that are relevant to the \mainword{} (i.e., marked as true by the evaluators). $\frac{\#\text{true events}} {\#\text{true events} + \#\text{false events}}$

\textbf{Relevance}: How relevant the timeline is to the word. This is meant to approximate the precision metric. Relevance was indicated as a rank score on a scale from 1 to $\#$timelines ($\#$timelines being the number of timelines presented for the given word) and normalized as: $\frac{\text{relevance score}}{\#\text{timelines}}$

\textbf{Missing Events}: Evaluators were asked how many events they believed were missing from the timeline. We then normalize by the total number of events in the timeline. Intuitively, this measure is meant to approximate \textit{1-recall}.
$\frac{\#\text{missing events}}{\#\text{events in timeline}}$

\textbf{Redundancy}: Evaluators were asked how many events in the timeline are redundant (normalized by the total number of events in the timeline):
$\frac{\#\text{redundant events}}{\#\text{events in timeline}}$

\textbf{Ranking}: Evaluators were asked to subjectively rank presented timelines from best to worst.

\textbf{Effectiveness}: Evaluators were asked to indicate their familiarity with the event history both before and after seeing the timeline. The difference between the two scores indicated the `effectiveness' (as a soft measure of whether the timeline contained anything surprising or novel). The pre-evaluation score also served to measure the evaluator's familiarity with the topic.

%\\This is meant to test whether evaluators feel that the timeline helped them gain a better understanding of the word and its history. %, which is, after all, our main goal.

\subsubsection*{Methods Compared}
We perform experiments for the two building blocks of a timeline. First, we compare methods of identifying turning points (Section~\ref{sec:year_detection}). We approximate a gold standard for turning points by having evaluators mark years in which there is a turning point. This is determined by looking at all the events that took place during a specific year and approximating whether any of them is significant to the \mainword{}. We compare the identified years of our methods to this gold standard.

Second, we evaluated different ways to produce descriptors\footnote{Refer to \url{https://github.com/guyrosin/generating_timelines} for the source code.}:

\textbf{WordDescriptors} (Section~\ref{sec:word_descriptors}): We use words as descriptors. For every year $t$, we select words that were added to the kNN of the \mainword{} during $t$ (with $k=20$, chosen empirically).

\textbf{BaseEvents}: %Close events to the \mainword{} can serve as good representatives of descriptors. %Specifically, we assume that an event influenced a word if the word appears many times in the event's page. 
A baseline method for event descriptors.
We select events `close' to the \mainword{} as descriptors---as measured by the frequency of the \mainword{}'s appearance on the event's Wikipedia page.

\textbf{WikiTimelines}: Finally, we extract timelines from crowd-created Wikipedia timeline pages\footnote{For example, \url{https://en.wikipedia.org/wiki/Timeline_of_Russian_history}}.

These baselines were compared to our algorithmic techniques: %Each one of them starts with detecting the relevant timepoints using the different embedding methods (Section~\ref{sec:year_detection}). We then identify events to use as descriptors.

\textbf{ByWord} and \textbf{ByKNN} are described in Section~\ref{sec:similarity_events}.

\textbf{ByKNNGlobCls}: We first use \textit{ByKNN} to find 30 close events (determined empirically), and then use the events classifier (Section~\ref{sec:classifier}) to predict each event's influence. We rank the events by a combination of this prediction and the score of \textit{ByKNN}. The classifier is trained on a dataset that was created using the global embeddings.

\textbf{ByKNNCls}: Similar to \textit{ByKNNGlobCls}, with one difference: here the classifier's training set was created using the projected embeddings.

\section{Results}
\label{sec:results}
%In this section, we outline the results of our empirical evaluation.

\subsection{Turning Point Evaluation}
\label{sec:timepoints_eval}
Comparing the two methods for turning point detection, we observed that 23\% of the years detected by \textit{Neighborhood} are false, compared to 15\% by \textit{EmbeddingSimilarity}.
We believe this difference is due to the \textit{Neighborhood} method capturing only the local neighborhood of the word, while an embedding can be more meaningful. For example, a word can change semantically not by altering its neighbors, but by moving in space towards other meanings, together with its neighbors.

\begin{table*}
  \centering
  \begin{tabular}{llllllll}
    \toprule
    Method & Accuracy & Relevance & Missing & Redundancy & Ranking & Effectiveness \\
    \midrule
    WordDescriptors & 0.43 & 0.32 & 0.65 & 0.3 & 0.28 & 0.03 \\
    BaseEvents & 0.49 & 0.76 & 0.04 & \textbf{0.03} & 0.72 & 0.23 \\
    WikiTimelines & \textbf{0.81} & 0.67 & \textbf{0.03} & \textbf{0.03} & 0.65 & 0.07 \\
    \midrule
    ByWord & 0.60 & 0.86 & 0.09 & 0.06 & 0.78 & \textbf{0.33} \\
    ByKNN & 0.61 & 0.81 & 0.12 & 0.08 & 0.80 & 0.31\\
    ByKNNGlobCls & 0.63 & 0.62 & 0.38 & 0.1 & 0.53 & 0.17\\
    \textbf{ByKNNCls} & 0.67 & \textbf{0.89} & 0.20 & 0.09 & \textbf{0.86} & \textbf{0.33}\\
    \bottomrule
  \end{tabular}
  \caption{Timelines evaluation results. Methods and metrics described in Section~\ref{sec:manual_timeline_evaluation}.}
  \label{tab:timelines_results}
%\vspace{-2mm}
\end{table*}

\subsection{Timeline Evaluation}
\label{sec:timeline_eval}
We present the results of the timeline evaluation (Table~\ref{tab:timelines_results}), where the turning points are detected using the \textit{EmbeddingSimilarity} method, as it reached the highest empirical performance (Section~\ref{sec:timepoints_eval}).

The \textit{WordDescriptors} method achieves poor results, as expected. It is inaccurate and contains many redundant descriptors. For example, it generated the following descriptors for the word ``Terror'' during 2012-2014: Islamic terror, brutality, genocidal. They are all related to terror but do not help us deduce why the word `Terror' was impacted, or what happened.
Alternatively, the \textit{ByWord} method performs much better. Looking at its generated timeline for `Terror', we observe that it successfully identifies highly relevant events: the Benghazi attack (a terror attack against US government facilities in Libya), the mass shooting at Westgate Shopping Mall in Kenya, and the international military intervention against ISIL (the Islamic State organization). 
We find that \textit{ByKNN} results and performance are similar to \textit{ByWord}. As both methods consider the similarity between an event and a \mainword{}, we conjecture that the similarity function has a less impact on the timeline generation process. 
%\textit{ByKNN} is an expansion of \textit{ByWord} in that it checks the similarity between an event and the \mainword{}'s neighbors. 
We empirically observe that in most cases a word close to an event would also be close to its neighbors.

The \textit{WikiTimelines} method has the top accuracy. Given these timelines are manually created by domain experts, this is unsurprising.
Nonetheless, the \textit{ByKNNCls} method wins the three most important metrics: relevance, ranking, and effectiveness. Thus in the eyes of the evaluators, this method gives the most relevant timelines compared to all others, and maybe most importantly, provides novel information. %and understanding of the history of the words.
As an example, the timeline created by \textit{ByKNNCls} for `Russia' contains several significant events that are missing from the same timeline created by \textit{WikiTimelines}, such as Chernobyl disaster, the Revolutions of 1989 and the Dissolution of the Soviet Union.

The \textit{ByKNNCls} method is more accurate than the other embedding-based methods, likely because it can filter out events that are identified by other methods but are in fact not impactful for the particular \mainword{}. However, it has a higher \textit{Missing Events} score---suggesting true events are occasionally filtered out as well.
%We can deduce that the \textit{ByKNNCls} method is the strongest overall.
%The differences between \textit{ByKNNCls} and its variant \textit{ByKNNGlobCls} are further discussed in Section~\ref{sec:projection_contribution_timelines}.

To measure the correspondence between evaluators' answers, we calculated Kendall's Tau, which resulted in an average value of 0.6.

\subsection{Events Classifier Evaluation}
\label{sec:classifier_results}
We experiment with several supervised approaches for the events classifier (Section~\ref{sec:classifier}) and evaluate their performance using stratified 10-fold cross-validation.
%Each classifier receives an event and a word, represented by the features described in Section~\ref{sec:classifier_features}, and outputs the probability this event affected the word. 
In this evaluation, the projected embeddings were used to create the training and test sets and for creating the classifier's features, since this configuration was found to result in the best performance (Section~\ref{sec:projection_contribution_classifier}).
Specifically, the parameters (optimized using grid search) and the AUC are as follows:

\textbf{Logistic regression} produced an AUC of 0.75.
\textbf{SVM} with RBF kernel and C=1.0 produced 0.97.
\textbf{Random Forest} classifier with 800 trees produced 0.97 as well.
\textbf{Neural Network} with a single hidden layer of 100 neurons and Adam as the optimization algorithm achieved the best performance, with an AUC score of 0.98.

\subsection{Projection Contribution}
We measure the embeddings projection's contribution (Section~\ref{sec:projected_embeddings}) to the tasks of timeline generation and learning influence of events on words.

\subsubsection*{Timeline Generation Performance}
\label{sec:projection_contribution_timelines}
We refer the reader to Table~\ref{tab:timelines_results} to discuss the comparison between \textit{ByKNNCls} and \textit{ByKNNGlobCls}. These methods are almost identical---both use our classifier for detecting significant events. They differ in how the classifier is trained. \textit{ByKNNGlobCls}'s training set is created using global embeddings, while \textit{ByKNNCls}'s training set is created using projected embeddings. We observe a significant difference in the performance of these two methods. %\textit{ByKNNGlobCls} is worse than the two other embedding-based methods (\textit{ByWord} and \textit{ByKNN}). It is slightly more accurate, but its other metrics are worse. We conjecture the reason for this is the performance of the classifier. %We discuss the classifier evaluation and the projection contribution to it in Section~\ref{sec:projection_contribution_classifier}. %The classifier causes many false negatives, i.e., true events that are not detected as such. As a result, its effectiveness score is significantly lower than of the other methods.
\textit{ByKNNCls} achieves the best performance of all methods. It has far fewer false negatives than \textit{ByKNNGlobCls} and higher accuracy. The other important metrics---relevance, ranking, and effectiveness---show improved performance as well. We conclude that representing events using the projected embedding brings high performance boosts for this task.

\subsubsection*{Events Classifier Performance}
\label{sec:projection_contribution_classifier}
To better understand the embedding features on the events classifier's performance (Section~\ref{sec:classifier}) we compared the impact of global and projected embeddings. % (note that the temporal word embeddings do not contain embeddings for events, so we could not use them here). 
Additionally, the training data of the classifier can be generated using any embedding. Thus, we perform an empirical evaluation comparing all combinations of embeddings for representing the \textit{features} and the \textit{training data} employed by the classifier (Table~\ref{tab:classifier_embeddings_comparison}).
Using global embeddings, we find an AUC of 0.69.
Using global embeddings for the features and projected embeddings for the dataset, or vice versa, resulted in AUC of around 0.84.
Using the projected embeddings for both yielded an AUC of 0.98---significantly better than any other combination---with $p<0.05$ (using a Wilcoxon signed-rank test).

\begin{table}
  \small
  \centering
  \begin{tabular}{llllll}
    \toprule
    Embeddings & Acc. & Rec. & Prec. & F1 & AUC\\
    \midrule
    Glob/Glob & 0.63 & 0.66 & 0.62 & 0.64 & 0.69 \\
    Glob/Proj & 0.74 & 0.73 & 0.74 & 0.74 & 0.81 \\
    Proj/Glob & 0.76 & 0.76 & 0.75 & 0.76 & 0.86\\
    \textbf{Proj/Proj} & \textbf{0.94} & \textbf{0.94} & \textbf{0.94} & \textbf{0.94} & \textbf{0.98}\\
    \bottomrule
  \end{tabular}
  \caption{Classifier performance using global and projected embeddings for creating its features/dataset.}
  \label{tab:classifier_embeddings_comparison}
%\vspace{-3mm}
\end{table}

\subsection{Events Classifier Contribution}
Our main goal in developing the classifier (Section~\ref{sec:classifier}) was to enable us to identify the relevant events that affect a given word. As presented in Table~\ref{tab:timelines_results}, the main drawback of \textit{ByWord} and \textit{ByKNN} is a high false positive ratio given this task. %, i.e., they detect many events that are actually not that significant for the \mainword{}.%, or that did not cause any change in its semantics.
The \textit{ByKNNCls} method is more accurate than the other embedding-based methods, probably due to the classifier filtering out many events that are identified by the other methods but are in fact not impactful for the particular \mainword{}. 
For example, we observe several false events that appear in the `Israel' timeline that was generated by the \textit{ByKNN} method and are all ignored by \textit{ByKNNCls}. Each one of them is related to Israel, but not significant enough to make an impact on it, e.g., the \textit{anthrax letters} (2001) had "death to Israel" written inside them. \textit{Hurricane Katrina} (2005) brought Israel to send a humanitarian aid delegation to New Orleans. 
Furthermore, the decrease in the false positive ratio results in high ratings of the \textit{ByKNNCls} timelines. Looking at the results in Table~\ref{tab:timelines_results}, we observe significant differences in multiple metrics: relevance, ranking, and effectiveness (tested using paired t-test with $p<0.05$). %We conclude that \textit{ByKNNCls} improves user knowledge and understanding of the history of the words.

\subsection{Discussion}
Three main factors seem to affect the performance of our approach.
First, ambiguity harms performance. For creating timelines for ambiguous words, a contextual embedding approach would be necessary. We leave that for future work. For example, ambiguous \mainword{}s such as `Oil' had worse scores than other similar words (e.g., `Tsunami' and `Disaster') and than expected. 
Second, a sufficient amount of significant events relevant to the \mainword{} is crucial, otherwise, the timelines would be too short in the eyes of the evaluators or the end users. For example, `ISIS' which is a relatively new organization, has a few significant relevant events and therefore had weak results in our evaluation.
Third, the available amount of data about the \mainword{} makes a difference. The more the better, as the temporal embeddings would then be rich and meaningful. We observed worse performance for relatively rare words, such as `Bombing', compared more common ones (e.g., `Attack' and `Russia').

% \begin{table}
%   \centering
%   \small
%   \begin{tabular}{lll}
%     \toprule
%     Year & Word Descriptors \\
%     \midrule
%     2001 & world trade center, quake, twin towers, grief \\
%     2005 & preparedness, katrina \\
%     2010 & spill, deepwater horizon, explosion \\
%     2011 & tsunamis, japan, fukushima, quake \\
%     2013 & typhoon, sandy, catastrophic \\
%     \bottomrule
%   \end{tabular}
%   \caption{A generated timeline with word descriptors for \textit{disaster}.}
%   \label{tab:disaster_timeline}
% \end{table}

% \begin{table}
%   \centering
%   \begin{tabular}{lll}
%     \toprule
%     Year & False Events \\
%     \midrule
%     2000 & Kursk submarine disaster \\
%     2001 & 2001 anthrax attacks \\
%     2002 & International Criminal Court Statute \\
%     2004 & 2004 Summer Olympics \\
%     2005 & Hurricane Katrina \\
%     2008 & Jeremiah Wright controversy \\
%     2011 & 2011 Christchurch earthquake \\
%     \bottomrule
%   \end{tabular}
%   \caption{False events that appear in the ``Israel'' timeline that was generated by the \textit{ByKNN} method, and are all ignored by \textit{ByKNNCls}.}
%   \label{tab:false_events_byknn}
% \end{table}

% \subsection{Qualitative Examples}

% \textit{Disaster}: The method detects both natural disasters and man-made disasters. hurricanes and tsunamis.

% \textit{Olympics}: The method detects every summer and winter Olympic games.

\section{Conclusions}
In this work, we develop methods to model the evolution of language in relation to world events. We introduced the task of timeline generation, which is composed of two components: identifying turning points when semantic changes occur, and representing descriptors (i.e., words or events in our case).
We presented several embeddings for the task and studied their effect. We find that our proposed method of projecting embeddings from a large, static model to a temporal one (i.e., from \textit{Wikipedia} to the \textit{New York Times}) yielded the best performance. Given several baselines we determined that a supervised approach leveraging the projected embeddings yields the best results. Using our method, high quality timeline generation can be done automatically and at scale.

\section*{Acknowledgements}
We thank Prof. Eytan Adar for early feedback and comments.

\bibliographystyle{acl_natbib}
\bibliography{00_main}
\end{document}